# Benchmarking Machine Learning: How Fast Can Your Algorithms Go?

Zeyu Ning, Hugues Nelson Iradukunda, Qingquan Zhang, Ting Zhu
*Department of Computer Science & Electrical Engineering*
*University of Maryland, Baltimore County*
Email: {zeyning1, hiraduk1, q, zt}@umbc.edu

*Abstract*—This paper is focused on evaluating the effect of some different techniques in machine learning speed-up, including vector caches, parallel execution, and so on. The following content will include some review of the previous approaches and our own experimental results.

*Index Terms*—machine learning, speed-up, parallel, cache

## I. Introduction

Since the beginning of this decade, machine learning has become the dominating methods in several different fields of application, such as computer vision, natural language processing, recommendation systems and search systems. One reason for machine learning methods' taking off in both academic and public domains is the abundant computation power which was not available in the past. Some techniques developed for graphic or scientific computations, have laid the foundation of present machine learning boom. The following sections will focus on evaluating some current and possible methods used in the speed-up of machine learning, especially neural networks, and we will propose some promising methods for future development.

## II. Speed-up Methods in General

The speed-up of algorithms is one of the fundamental problems in computer architectures, the optimization of computer architectures for better performance has never stopped, and significant achievements have been made by introducing RISC [1] in 1983 and the trend of multi-processor around 2005. Besides these major changes, there are many techniques introduced during the period that have improved the performance of computers significantly, including dynamic scheduling [2], branch prediction [3], vector cache [4] and SIMD architecture [5]. Not only have these techniques improved the performance, but also inspired hardwares with specific applications. One of the most known hardwares for specific applications is GPUs or Graphic Processing Units. These processors have substantial number of threads and large register files; they were initially designed for the complex matrix operations in graphics, but recently they are playing a critical role in machine learning and deep learning due to the ability to handle large scale data-parallel processing.

## III. Speed-up Methods in Different Hardwares

Modern computers are a mixture of heterogeneous hardwares, and as transistors are reaching their physical size limits, there'll be more specific components for specific tasks in the future. As a consequence, the different hardware architectures would use different scaling techniques according to their specific tasks.

One simple example is multi-processors vs. single-processor. Usual multi-processor will have multiple threads; the number of threads may be the same as processor, maybe not. Thus the multi-processor system introduced thread-level parallelism: each thread will have its own instructions and data, and the communication between threads are reduced to the minimal degree, this scheme works well in modern computers when multi-tasking, however for each processor, it didn't improve the performance for each task by a significant factor. While for single-processor or each processor in multi-processor, usually instruction-level parallelism techniques such as loop unrolling [7] or branch prediction [7] will be used to improve performance.

However, instruction-level parallelism (ILP) is ultimately dependent on the multiple issuing, but increasing issuing rate will reduce the maximum clock rate [7]. So instead of ILP, data-level parallelism is the way for speeding up used by GPUs, specifically for their main task of vector operations.

The following graph [8] has shown the architecture of Nvidia GPUs; From the graph above, we can clearly tell the difference between GPUs and CPUs: the larger number of functional units, large shared cache in GPU have made it possible for data-level parallelism in scale, and for some specific tasks this architecture can reach a speed-up up to 130x [8].

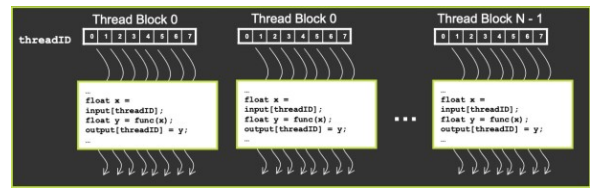

Fig. 1. Blocks and threads in Nvidia GPUs: shared cache within each block

## IV. Speed-up Methods in Machine Learning

Machine learning is one of the major topics going on in current computer science, the contents of machine learning has included many techniques ranging from simple to complex, from linear regression to multi-layer perceptron, and deep neural networks. The speed-up methods we mentioned previously may work for some cases, but not all of them.

Taking multi-layer perceptrons as an example, for each training instance we input, the forward propagation can be executed independently without any data-sharing, but for each data dimension in the training set, since the full-connection layer would have data dependency on each other, explicit thread-level parallelism is not feasible.

Currently, the major speed-up method used in most machine learning algorithms is utilizing the data-level parallelism or thread-level parallelism [9], and with the help of SIMD [5] and vector instruction sets, these parallelisms are easier to achieve today. One noted example is MapReduce [10], it's a data-level parallelism architecture which supports a series of machine learning algorithms, including neural networks [11], k-means clustering [12] and logistic regression [13]. The following experiment will focus on the application of similar approaches in neural networks.

Another important method in the speed-up of neural networks is batch gradient descent [14], this architecture has eliminated the need to update the weights after each training instance (the online training fashion), thus for each sample instance in the same batch, the same weights will be used to calculate the output. Thus, the training can be finished in a SIMD fashion. Usually, the implantation of this fashion is using a vector instruction set [15], however in the following experiments we will explore a new method in speed-up.

## V. Experiment Methodology & Performance Criteria

In the following experiments we are testing the effectiveness of the current speed-up methods used in neural networks, as well as comparing our raw implantation with the implantation in current deep learning frameworks (Pytorch, Keras). The baseline model will have the same structure as all the implantations mentioned, but it's in a fully sequential execution fashion, i.e. online gradient descent for each sample (mini-batch size is 1), sequential weight updating, and so on.

The performance for each implantation will be evaluated based on the time of execution certain number of iterations, and the speed-up factor for each implantation. Also, we will evaluate the speed-up of each sub-operation in the whole training process, thus evaluate their influence on the overall performance.

According to the part in which our implantation can speed-up, we will derive the theoretical speed-up factor by Amdahl's law:

$$S(s) = 1/((1 - p) + p/s)$$

The structure of the testing neural network is shown in the following graph:

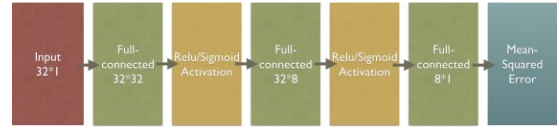

Fig. 2. Neural network structure

Also, besides the speed-up using parallel computation, we will also analyze the effect of different design choices in deep learning neural networks, for example, the choice of activation functions and loss functions. The design choices in the deep learning neural networks can be described in the following graph:

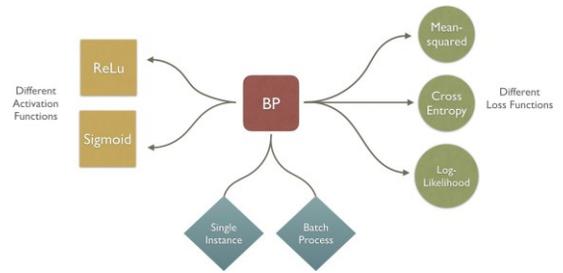

Fig. 3. Neural network design choices

## VI. Experiment Results & Analysis

- Comparison between ReLu & Sigmoid activation functions

Relu and Sigmoid functions are two most frequently used activation functions in deep neural networks, in which ReLu is the most popular one due to its better gradient properties. The following experiments have finished 1000 epochs in each configuration, and we have repeated it 4 times to ensure the consistency of the timing, the time used is shown in Fig. 4: From the graph we can find that in the raw implantation of neural network shown in Fig.2, ReLu activation function is significantly slower than Sigmoid function. The reason for this might be due to the implantation of condition clauses in Python, since the ReLu activation function would involve condition clauses when determining the input is less or more than 0. In order to further evaluate the possible cause of this, we split the process into front-prop and back-prop so that we can evaluate the effect of each operation on the final outcome, shown in Fig. 5 and Fig. 6: The experiments above are also finished in the same fashion: 100 epoches and 4 times each experiment; but the results are interesting since the front-prop of Sigmoid is really slow, while the back-prop for ReLu is slower than Sigmoid. The overall result is caused by the slow front-prop for ReLu. In order to look into this matter in detail, we looked at the raw implantation and found that the ReLu will run two conditional clauses in sequence, thus floating-point calculation in Python is less than conditional clauses.

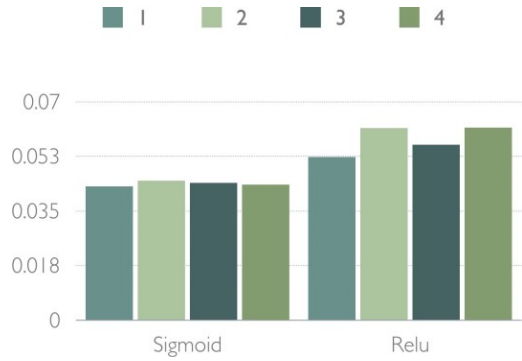

Fig. 4. ReLu vs. Sigmoid in raw implantation

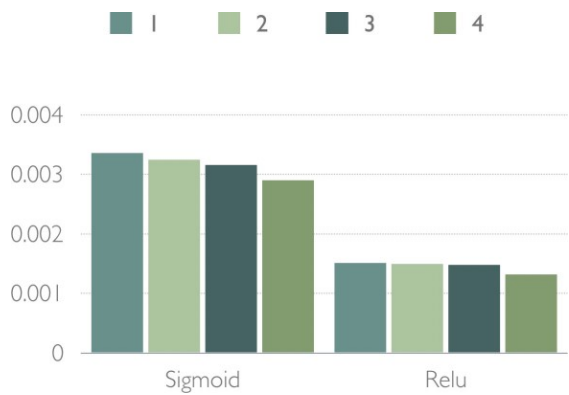

Fig. 5. ReLu vs. Sigmoid when front-prop

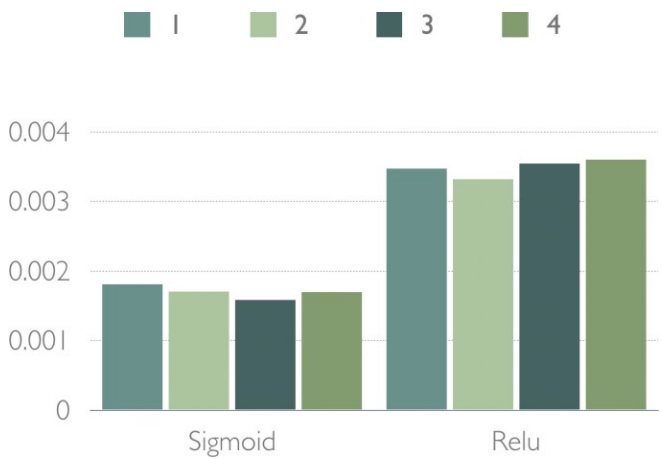

Fig. 6. ReLu vs. Sigmoid when back-prop

- ReLu & Sigmoid activation functions in frameworks
  We have also explored the performance of these activation functions in deep learning frameworks, Pytorch in the following example. The result is shown in the following graph:

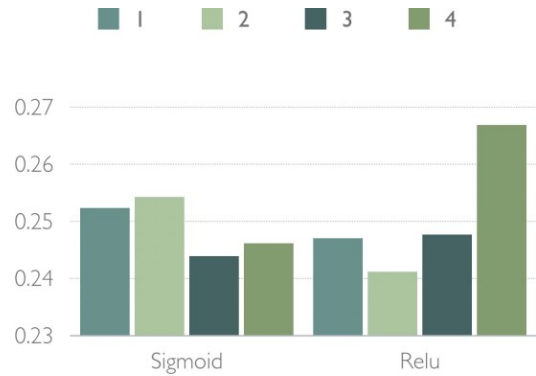

Fig. 7. ReLu vs. Sigmoid in Pytorch

From the graph above, there're two significant insights we can achieve:

1) Using frameworks are generally really slow compared to raw implantation. A key reason for it is that since the frameworks would have to consider its flexibility between different neural network structure, while raw implantations are just for this structure, frameworks will use a lot more variables to hold some values for potential use in the future, thus the cost of visiting memory is way larger than raw implantations. As a result, the overhead introduced is huge.

2) The difference in ReLu and Sigmoid function execution is minimal. Since the overhead introduced is already 10 times higher than the operations themselves, thus the difference in execution time is mainly covered by the overhead. This justifies the choice of ReLu function in real life, since usually ReLu function converges in less epochs, if ReLu is not significant slower than Sigmoid in each epoch (about 1.325x time in Fig. 4), ReLu can make training faster to converge.

Using a different framework will not change the statement we made before, in Keras the two functions still have comparable speed.

Also, we made a comparison of raw implantation and frameworks in the following Fig. 9, clearly raw implantation is way faster than others, while Keras, as a meta-framework based on Tensorflow, it's the slowest one.

- Data-level Parallelism vs. Batch Gradient Descent

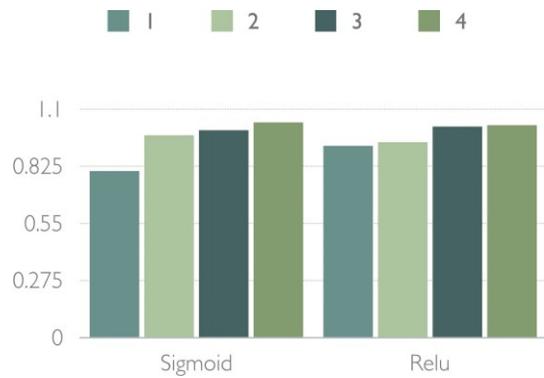

Fig. 8. ReLu vs. Sigmoid in Keras

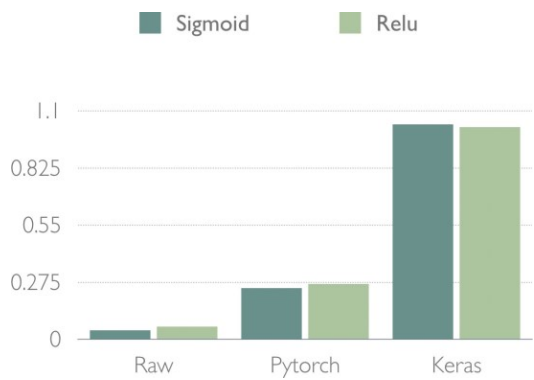

Fig. 9. ReLu vs. Sigmoid in raw implantation, Pytorch and Keras

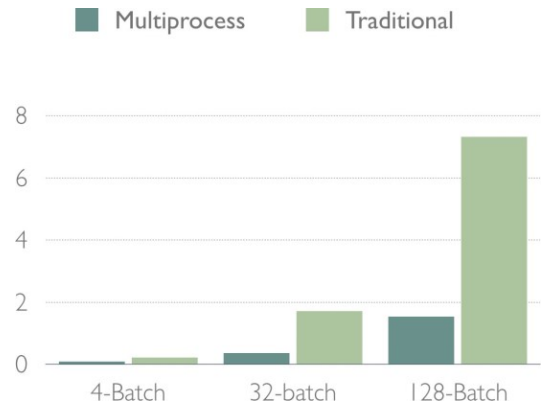

Fig. 10. Comparison multi-thread situation and baseline

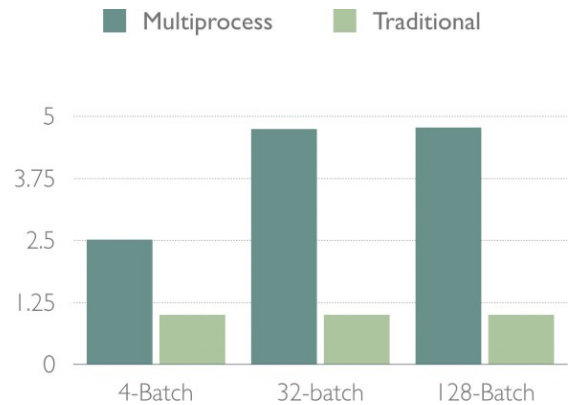

Fig. 11. The speedup that multithread can achieve compared to baseline

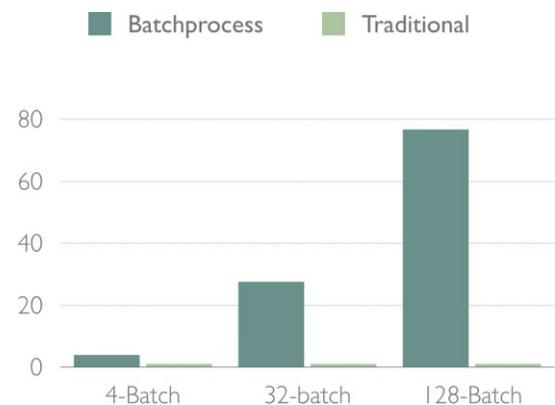

Fig. 12. The speedup that batch processing can achieve compared to baseline

The next experiment we did is to compare two different ways to achieve data-level parallelism. Batch gradient descent is stacking the N training instances onto each other to make an NxM vector, while another way we did is to spawn an additional thread for each new training instances. At the bottom, these two methods are the same, since the GPU is eventually using multiple threads to handle the NxM vector calculation. However, due to the limit of computation power, we can only use a CPU-based multi-thread solution to achieve the outcome, thus the theoretical speedup cannot exceed 6, since there're only 6 cores in the intel i7-8850M processor that we are using.

The following graph Fig. 10 shown the difference between the runtimes, when we spawn a new thread for each new data instance and the baseline model when we don't.

From the Fig. 10 we can see that the maximum speedup is x4, smaller than the x6 that we anticipated. So according to the Amdahl's law, the percentage that we can have a x6 speedup is about 90%.

Now let's look at the effect of batch processing situation in Fig. 12. Clearly, batch processing can achieve a much larger speedup than the multithread case. However, in the multithread case our overall speedup is limited by the number of CPU cores, while the limit to batch processing is only the number of multipliers in the GPU (if the NumPy package in Python has utilized the GPU) and the size of vector caches.

In order to evaluate the effect of vector cache size on the overall speedup performance, we have changed the batch size multiple times and trying to find the maximum limit of the speedup. According to the vector cache properties, the speedup times cannot exceed the maximum length of the vector cache, given only one layer of vector cache is used; Since if we have to fit another portion of the batch into the batch, we will have to load the cache again, this would make it slower. Thus in the following Fig. 13 we chose different batch sizes as large as more than 1024, the speed is shown in the following graph. If they are all within the maximum vector batch length, the graph should be linear.

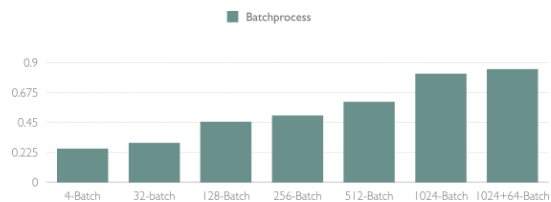

Fig. 13. The runtime of different batch size batch processing

According to the graph, most of them are linear while from 32-batch to 128-batch, the increase is less than 4 times, thus we can suppose that there're more than 1 level of cache, and the first layer of cache has a length around 64.

## VII. RELATED WORK

In the area of machine learning, researchers used Recurrent Neural Networks (RNNs) models, such as PL, GENPass [16]. These models were proposed to optimize both matching rates in passwords guessing and generality efficiency.

Researchers proposed various approaches to optimize performance of wireless signals systems [48-50] and IoT network devices [31-44]. Harmony [17] is one approach designed to optimize the accuracy of human activity recognition and monitoring services. The effectiveness of Harmony was evaluated by measuring the Received Signal Strength (RSS) values among the IoT devices. Another design used for robust fingerprinting, AP-Sequence [18] was proposed to optimize localization due to its ability to handle the dynamic power control.

Recent advances in the smart grid [19-30] and smart health [45-47] used optimization methods for improving the energy consumption at a large scale in the grid.

The work presented in this paper emphasizes the difference and effectiveness of the speed-up algorithms and hardware components involved in machine learning optimization.

## VIII. CONCLUSION & PROSPECT

In the previous parts, we have explored the existing methods used in machine learning, especially in neural networks. Our experiment has proved that data-level parallelism is useful for neural networks speedup, and the limit to this method is the number of cores or threads in CPU or GPU.

Due to the time constraints, we didn't compare the effectiveness of our multi-thread implantation and the one used in MapReduce paper [10]. There's a slight difference in these two methods, our method spawns a new thread for each new data piece, this thread will run a complete process while each individual thread in the MapReduce paper will only run the gradient descent part. This part is very important, and it's expected to finish in the near future.